\documentclass[10pt,twocolumn,letterpaper]{article}

\usepackage{cvpr}
\usepackage{times}
\usepackage{epsfig}
\usepackage{graphicx}
\usepackage{amsmath}
\usepackage{amssymb}

\usepackage[ruled,vlined]{algorithm2e}

\usepackage{times}
\usepackage{epsfig}
\usepackage{graphicx}
\usepackage{color}
\usepackage{epsfig}
\usepackage{comment}
\usepackage{enumitem}
\usepackage[export]{adjustbox}
\usepackage{tabularx}
\usepackage[table]{xcolor}
\usepackage{colortbl}
\usepackage{subfig}
\usepackage{multirow}
\usepackage{booktabs}
\usepackage{xr}
\usepackage{pifont}
\usepackage[hyphens]{url}
\usepackage{sidecap}

\usepackage{amssymb}
\usepackage{amsmath}

\usepackage{ragged2e}
\newcommand{\cmark}{\ding{51}}%
\newcommand{\xmark}{\ding{55}}%

\usepackage[pagebackref=true,breaklinks=true,letterpaper=true,colorlinks,bookmarks=false]{hyperref}

\cvprfinalcopy 

\def\cvprPaperID{4508} 

\ifcvprfinal\fi
\begin{document}

\title{Learning Nonparametric Human Mesh Reconstruction from a Single Image without Ground Truth Meshes}

\author{Kevin Lin\textsuperscript{\dag} \ \ Lijuan Wang\textsuperscript{\ddag} \ \ Ying Jin\textsuperscript{\dag\ddag} \ \  Zicheng Liu\textsuperscript{\ddag} \ \ Ming-Ting Sun\textsuperscript{\dag}\\
\textsuperscript{\dag}University of Washington \ \ \ \ 
\textsuperscript{\ddag}Microsoft\\
{\tt\footnotesize \{kvlin,mts\}@uw.edu,  \{lijuanw,Ying.Jin,zliu\}@microsoft.com}
}

\maketitle

\begin{abstract}

Nonparametric approaches have shown promising results on reconstructing 3D human mesh from a single monocular image. Unlike previous approaches that use a parametric human model like skinned multi-person linear model (SMPL), and attempt to regress the model parameters, nonparametric approaches relax the heavy reliance on the parametric space. However, existing nonparametric methods require ground truth meshes as their regression target for each vertex, and obtaining ground truth mesh labels is very expensive. In this paper, we propose a novel approach to learn human mesh reconstruction without any ground truth meshes. This is made possible by introducing two new terms into the loss function of a graph convolutional neural network (Graph CNN). The first term is the Laplacian prior that acts as a regularizer on the reconstructed mesh. The second term is the part segmentation loss that forces the projected region of the reconstructed mesh to match the part segmentation. Experimental results on multiple public datasets show that without using 3D ground truth meshes, the proposed approach outperforms the previous state-of-the-art approaches that require ground truth meshes for training. 

\end{abstract}

\section{Introduction}\label{sec:intro}

Estimating the 3D shape of a human body is one of the fundamental challenges in computer vision. Human mesh reconstruction~\cite{loper2015smpl, weng2019photo, Guler2018DensePose, ionescu2014human3,lassner2017unite,pavlakos2018learning,Guler2018DensePose, kolotouros2019convolutional,vonMarcard2018,kanazawa2019learning} has recently drawn an increasing attention as it plays an important role for a variety of applications such as augmented reality, human-computer interaction, and activity analysis.
While many studies have demonstrated effective 3D reconstruction using depth sensors~\cite{newcombe2011kinectfusion,shin20193d}, inertial measurement units (IMUs)~\cite{zheng2018hybridfusion,DIP:SIGGRAPHAsia:2018,vonMarcard2018}, and multiple cameras~\cite{natsume2019siclope,joo2018total,Pavlakos_2019_ICCV}\textcolor{black}{, people are exploring to use a monocular camera setting which is more convenient and efficient. However, it remains} challenging to reconstruct human mesh from a single monocular image due to complex deformation of the human body, object occlusion, and limited 3D information.

\begin{figure}
\setlength{\tabcolsep}{0.5pt}
	\centering
	\includegraphics[width=0.32\columnwidth]{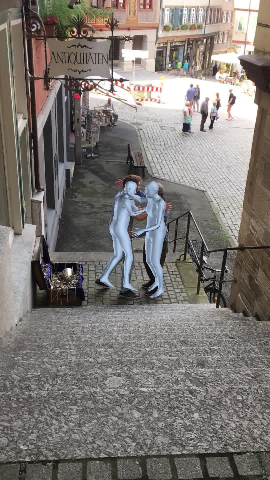}
	\includegraphics[width=0.32\columnwidth]{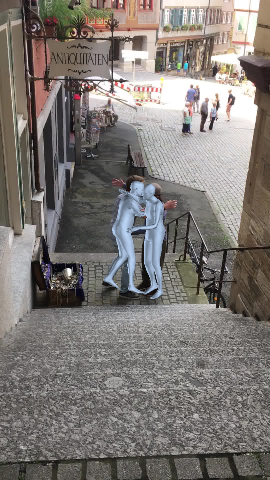}
	\includegraphics[width=0.32\columnwidth]{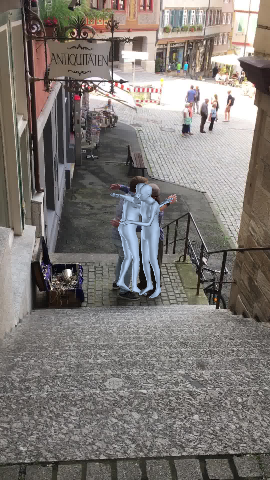}\\
	\captionof{figure}{
	Our nonparametric approach reconstructs human mesh without ground truth mesh training labels, and performs more favorably against the previous state-of-the-art nonparametric methods that use mesh training labels.
	}
	\label{fig:teaser}
\end{figure}

Supervised training with deep convolutional neural networks have shown great progress on human mesh reconstruction from a single image. However, many existing approaches~\cite{lassner2017unite,pavlakos2018learning,Guler2018DensePose,kolotouros2019convolutional,zanfir2018monocular} require ground truth mesh labels for training. Since it is difficult and expensive to capture ground truth meshes for a large variety of scenes, it is desirable to avoid the requirement on the ground truth meshes.  To address the problem, recent studies~\cite{pavlakos2018learning,kanazawa2018end,omran2018neural,guler2019holopose} propose to use a parametric human model such as skinned multi-person linear model (SMPL)~\cite{loper2015smpl} and regress the shape and pose coefficients. Great success has been achieved by using the parametric human model. However, parameter
regression remains a challenging task and it usually requires a large number of paired image-SMPL data for supervised training. On the other hand, the parametric representation has limitations. Construction of the model like SMPL requires digitizing a large number of people with different shapes and poses, and it is time consuming and expensive. In practice, only a limited amount of shape and pose variations can be captured in a dataset, and as a result, the resulting parameter space may not cover all the variations in the real world.

\begin{figure*}[t]
	\centering
    \includegraphics[width=0.85\textwidth]{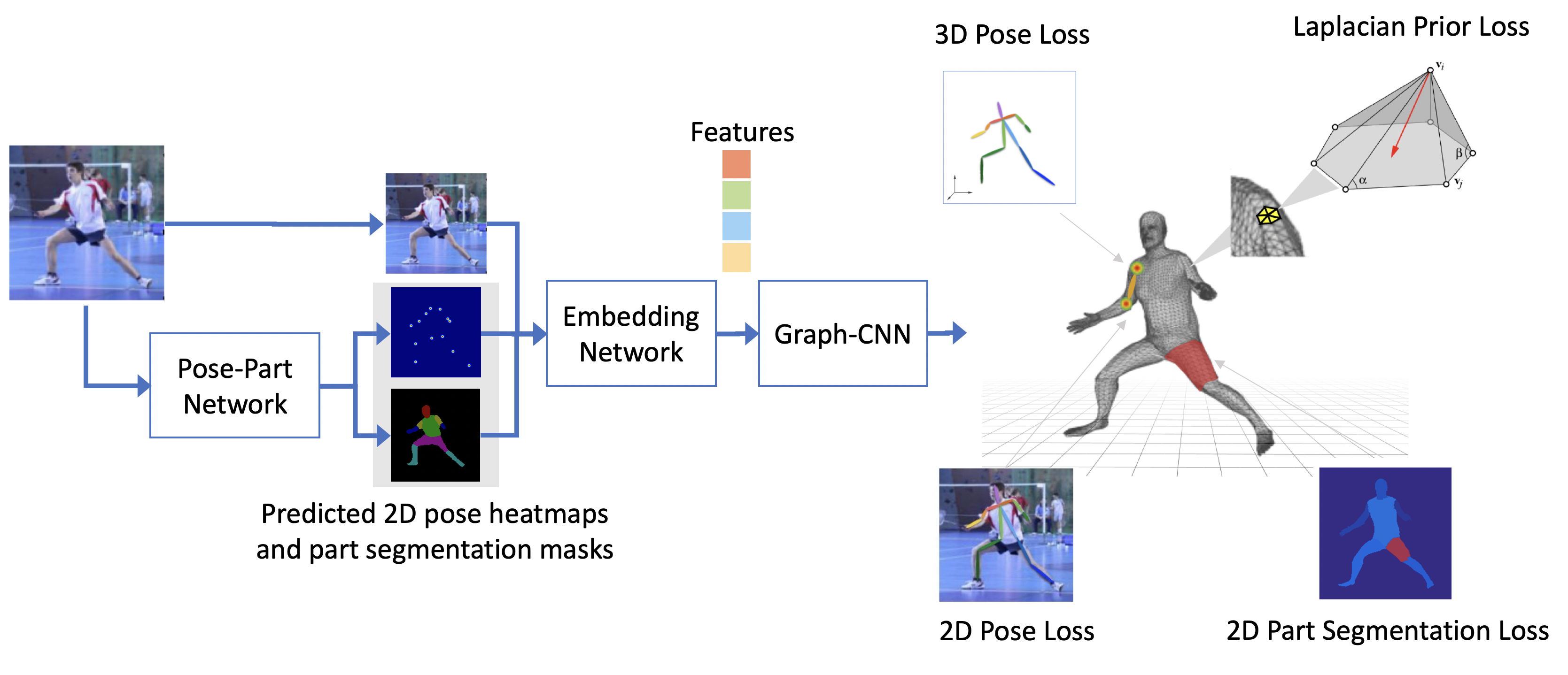}
	\captionof{figure}{
	An overview of our mesh reconstruction framework. It consists of three subnetworks: (1) Pose-Part Network 
	that extracts the pose heatmaps and part segmentation masks from the input image, (2) Feature Embedding Network that takes as input the image and the pose and part segmentation heatmaps and outputs a feature vector for the Graph CNN, and (3) Graph Convolutional Neural Network that takes as input the feature embedding and outputs the 3D coordinates of all the mesh vertices.}
	\label{fig:first}
\end{figure*}

In this paper, we propose a method that does not require ground truth meshes for training and does not regress the parameters of a parametric human model. Like the recent state-of-the-art approach~\cite{kolotouros2019convolutional}, we represent human mesh in a form of \textit{graph}, and use a graph convolutional neural network (Graph CNN) to learn human mesh reconstruction. Since we do not use ground truth meshes in training, we introduce two new terms in the loss function. The first term is the Laplacian prior that acts as a regularizer on the reconstructed mesh. Laplacian prior has been used widely for geometric modeling and mesh editing~\cite{taubin1995signal,sorkine2004laplacian,nealen2006laplacian}, but we are the first to use it with Graph CNN to learn mesh reconstruction. The second term is the part segmentation loss that forces the projected region of the reconstructed mesh to match the part segmentation. Since the existing datasets like UP-3D~\cite{lassner2017unite} and Human3.6M~\cite{ionescu2014human3} do not contain many scenes with occlusions, the learned model usually does not handle occlusions very well. To address this problem, we propose to feed 2D pose and part segmentation heatmaps to the feature embedding network. The 2D pose and part segmentation heatmaps are learned by leveraging the existing pose and part segmentation datasets like MSCOCO~\cite{lin2014microsoft} and Pascal-Person-Parts~\cite{chen2014detect} both containing large number of images with occlusions. As a result, our method works much better in handling occlusions.

In summary, the main contributions of this paper include:

\begin{itemize}




\item We are the first to learn nonparametric body shape reconstruction from a single image without mesh supervision.
\item Experimental results on multiple datasets show that our proposed method achieves comparable or better performance than the state-of-the-art methods which require ground truth meshes in training.
\item  By explicitly feeding 2D pose and part segmentation heatmaps into the feature embedding network, the robustness in occlusion scenarios has been significantly improved, as demonstrated in the qualitative comparison results in Section 4.4.


\end{itemize}







\section{Related Works}\label{sec:related}

\subsection{Parametric regression}

Human mesh reconstruction using parametric approaches has a long-standing history. Majority of the previous works adopt the
SMPL parametric model~\cite{loper2015smpl}, and propose to regress the shape and pose parameters. The regression can be done with the help of various 2D human body features such as human skeletons~\cite{lassner2017unite,pavlakos2018learning}, silhouettes~\cite{pavlakos2018learning}, and body part segmentations~\cite{omran2018neural}. Kanazawa~\textit{et al.}~\cite{kanazawa2018end} proposed to integrate the differentiable SMPL model as a layer within a neural network, and estimate SMPL parameters from an input image using pose prior with an adversarial training framework. Tung~\textit{et al.}~\cite{tung2017self} proposed to learn SMPL parameters using a self-supervised strategy.

\subsection{Nonparametric body shape estimation}

Instead of using a parametric model, various work has been reported to directly estimate the body shape from an image including leveraging depth~\cite{shin20193d}, or 2D-to-3D correspondences~\cite{Guler2018DensePose}, and representing 3D mesh into a volumetric space~\cite{varol2018bodynet} or graph~\cite{kolotouros2019convolutional}. 
Specifically, Varol~\textit{et al.}~\cite{varol2018bodynet} proposed to embed the 3D mesh into a volumetric space for learning human body shape. The volumetric representation is memory intensive resulting in a limited resolution. In addition, it requires ground truth meshes obtained from synthetic dataset.
G{\"u}ler~\textit{et al.}~\cite{Guler2018DensePose,alp2017densereg} proposed to associate image pixels with part-based UV maps. However, manual label acquisition is very expensive, and model prediction does not explicitly provide semantic information of the 3D geometry. 
Kolotouros~\textit{et al.}~\cite{kolotouros2019convolutional} showed that the regression can be significantly easier than the conventional approaches by using graph convolutional neural networks (Graph CNNs), but it requires well-annotated ground truth meshes since its regression target for each vertex is its 3D ground truth location.
Zhu~\textit{et al.}~\cite{zhu2019detailed} proposed a multi-stage deformation refinement, and used depth information to find surface variation. But it needs to manually define the handles on the surface for controlling the mesh deformation. 
Natsume~\textit{et al.}~\cite{natsume2019siclope} cast the problem as a multi-view silhouette-based reconstruction, but rely heavily on multi-view segmentation synthesis. Saito~\textit{et al.}~\cite{saito2019pifu} proposed to learn a 3D occupancy field using depth information. Among the literature, the common theme of all these works is that they have focused on strongly supervised learning using labeled training data. However, the acquisition of large-scale 3D mesh labels, especially for human body shape, is very expensive. We propose to relieve the need for ground truth meshes by formulating a new learning objective function using Laplacian prior and part segmentation in a Graph CNN framework.

\subsection{Graph convolutional neural networks for vision}
While deep convolutional neural networks~\cite{lecun2015deep} are effective for extracting hidden patterns from data, there are many computer vision tasks where the data can be represented in a form of graph~\cite{wu2019comprehensive}. By using the graph as the representation, it is shown to be more effective for high-level semantic analysis, such as scene graph generation~\cite{xu2017scene,yang2018graph}, image content generation~\cite{johnson2018image}, category-specific object modeling~\cite{wang2018pixel2mesh}, 3D hand estimation~\cite{ge20193d, zhang2019end, wan2019self}, face reconstruction~\cite{ranjan2018generating}, human action recognition~\cite{jain2016structural,yan2018spatial}, human-object interaction~\cite{qi2018learning}, semantic segmentation~\cite{qi20173d}, and image classification~\cite{garcia2017few}. Graph CNN has been recently used~\cite{kolotouros2019convolutional,litany2018deformable,verma2018feastnet} to estimate the 3D shape of a human body, however, these methods require the 3D ground truth locations for each vertex of the mesh as their regression target. These limitations have motivated us to develop a technique that does not require ground truth mesh supervision.

\section{Method}\label{sec:method}

Figure~\ref{fig:first} is an overview of our framework. Like~\cite{kolotouros2019convolutional}, we also use a graph CNN, but there are three main differences. First, we do not use ground truth meshes, and thus we do not have the 3D mesh vertex loss term. We instead add a Laplacian prior term which regularizes the 3D mesh reconstruction. Second, we add a Pose-Part network and feed the extracted pose heatmaps and part segmentation masks to the feature embedding. Third, we add a part segmentation loss term which ensures that the reconstructed shape is consistent with the projected region for each part of the body.



\subsection{Image-based feature embedding}

In the first part of our model, we use a Pose-Part Network similar to the existing multi-task networks~\cite{he2017mask,lin2019cross} to predict pose heatmaps and part segmentation masks from the input image. We concatenate the input image and the human-related feature maps, and use a CNN to extract a feature embedding. In this work, we use a ResNet50~\cite{he2016deep} to extract feature embedding.

Assume we have a dataset $D$ with 2D pose labels, 3D pose labels, and 2D part segmentation labels. Let $D=\{I^i, \bar{J}_{2D}^i, \bar{J}_{3D}^i,\bar{B}_{2D}^i\}_{i=1}^{H}$, where $H$ is the total number of training images, $I \in R^{w\times h \times 3}$ denotes an image, $\bar{J}_{2D} \in R^{K \times 2}$ denotes the ground truth 2D coordinates of the joints and $K$ is the number of joints on a person.
Similarly, $\bar{J}_{3D} \in R^{K \times 3}$ denotes a 3D joint ground truth. 
$\bar{B}_{2D} \in R^{w\times h \times Z}$ is the body part segmentation ground truth and $Z$ is the total number of body part categories. We first train our Pose-Part Network using $\bar{J}_{2D}$ and $\bar{B}_{2D}$. Then, we train our entire model using $\bar{J}_{2D}$, $\bar{B}_{2D}$, $\bar{J}_{3D}$ to reconstruct human mesh. 
\subsection{Graph CNN}

The Graph CNN in our proposed method reconstructs human mesh by applying the projection matrix to the input feature vectors $X$ and then compute the vertex coordinates:
\begin{equation}
\label{eqn:abstract}
Y = F(X;\bar{A},W),
\end{equation}
where $\bar{A} \in R^{N\times N}$ denotes the adjacency matrix of the human mesh. $X \in R^{N\times d}$ denotes a set of $d$-dimension feature vectors which are the output of the embedding network. $Y \in R^{N\times 3}$ is the estimated 3D coordinate for the mesh vertices.  $F(X;\bar{A},W)$ is a composition of a number of projections which can be written as:
\begin{equation}\label{eqn:dnn}
	F(X;\bar{A},W) = f_T( \cdots f_{2}(f_{1}(X;\bar{A},W_{1});\bar{A},W_{2})\cdots;\bar{A},W_{T}),
\end{equation}
where $f_t(\cdot)$ takes the input $X_t$, the adjacency matrix $\bar{A}$, and parameter $W_t$ as inputs, and
produces the projection result $X_{t+1}$ by using:
\begin{equation}\label{eqn:dnn}
	X_{t+1} = f_{t}(X_t;\bar{A},W_{t})=\sigma(\bar{A}X_tW_t),
\end{equation}
\textcolor{black}{where $\sigma(\cdot)$ is the activation function introducing non-linearity to the network model. We use rectified linear unit (ReLU) in this work.}

The proposed method aims to learn a series of Graph Convolution layers, which are $T$ projection matrices $W = \{W_1, W_2, \cdots, W_T\}$ that map the input feature vectors $X$ into the output vertex coordinates $Y$. We use the following objective function to learn $W$:
\begin{equation}
\textcolor[rgb]{0.00,0.00,0.00}{
\begin{aligned}\label{eqn:overall-loss}
\min_{W}\mathcal{L}(W) & =  \mathcal{L}_{Lap}(W) + \mathcal{L}_{3DPose}(W)\\ & + \mathcal{L}_{2DPose}(W) + \mathcal{L}_{2DPart}(W),
\end{aligned}
}
\end{equation}
where $\mathcal{L}_{Lap}$ is the Laplacian prior term, $\mathcal{L}_{3DPose}$ is the 3D pose loss, $\mathcal{L}_{2DPose}$ is the 2D pose loss, and $\mathcal{L}_{2DPart}$ is the part segmentation loss. We elaborate these loss terms in more detail in the following.


\subsection{Laplacian prior}

Laplacian prior has been commonly used for geometric modeling and mesh editing~\cite{sorkine2004laplacian,zhou2005large,nealen2006laplacian,desbrun1999implicit}. In this work, we are the first to use it with Graph CNN to learn human mesh reconstruction. Let $G$ denote the 3D mesh of a generic human body, where $G$ can be represented as a graph $G=(V,E)$ with the edges $E$ and the vertices $V$. We denote $V=[v_1, v_2, \cdots, v_n]$ and $v_i=[v_{ix}, v_{iy}, v_{iz}]$. Given a vertex $v_i$, the Laplacian of $v_i$ can be written as:\begin{equation}
\label{eqn:laplacian}
\delta_i = \sum_{\{i,j\}\in E} w_{ij}(v_{i}-v_{j}) = v_{i}  - \left[ \sum_{\{i,j\}\in E} w_{ij}v_{j}\right],
\end{equation}
where $\sum_{\{i,j\}\in E} w_{ij}=1$. To compute the Laplacians for the human body mesh, assume we have $n$ vertices in the mesh, which means $V = [v_1, v_2, \cdots, v_n]^T$. We can use a $n \times n$ Laplacian matrix:
\begin{equation}
\label{eqn:laplacianmatrix}
L_{i,j} =
  \begin{cases}
    w_{ij}       & \quad \text{if } \{i,j\}\in E\\
    -1  & \quad \text{if } i=j\\
    0 & \quad \text{otherwise},
  \end{cases}
\end{equation}
and compute the Laplacians $\Delta=[\delta_1,\delta_2,\cdot,\delta_n]^T$ using
\begin{equation}
\label{eqn:laplacianmatrix}
\Delta = LV.
\end{equation}

In the reminder of this paper, we use the uniform Laplacian~\cite{desbrun1999implicit,nealen2006laplacian} where the 1-ring vertex neighbors are equally weighted. The uniform Laplacian of $v_{i}$ points to the centroid of its neighboring vertices, and has the nice property that its weights do not depend on the vertex positions. To obtain the Laplacian for the entire mesh, we compute the $x$, $y$ and $z$ coordinates of the Laplacian $\Delta_{d}=\lbrack\delta_{1d},\delta_{2d},...,\delta_{nd}\rbrack^{T}, d \in \left\{ x, y, z \right\}$,
separately as 
\begin{equation}
\begin{split}
\Delta_{d}=L V_{d}.
\end{split}
\end{equation}

Unlike a rigid object mesh where its Laplacian is a constant, the shape of human body can be deformed in various ways depending on different poses and body movements. To learn the Laplacian prior, we randomly
sample pose parameters and generate a large number
of meshes with different poses for the average person
in the SMPL database~\cite{loper2015smpl}. We model the density distribution of Laplacian under the framework of Gaussian Mixture Model (GMM). Given a batch of $M$ mesh samples, we can estimate the parameters in GMM as follows.\begin{equation}
\begin{split}
P(\Delta_{d})=\sum^K_{k=1}  \hat{\phi}_{dk}\mathcal{N}(\Delta_{d}| \hat{\mu}_{dk}, \hat{\Sigma}_{dk}),
\end{split}
\end{equation}
where $\hat{\phi}_{dk}, \hat{\mu}_{dk}, \hat{\Sigma}_{dk}$ are mixture probability, mean, co-variance for component $k$ in GMM for $\Delta_{d}$, $d \in \left\{ x, y, z \right\}$ respectively. With the estimated parameters, the overall loss function for the Laplacian prior is written as:
\begin{flushleft}
$\mathcal{L}_{Lap}(W)=$
\end{flushleft}
\begin{equation}
\begin{aligned}
\sum_{d \in \left\{ x, y, z \right\}} -\log \sum_{k=1}^K \hat{\phi}_{dk}
 \frac{\exp\left(-\frac{1}{2}(\Delta_{d}-\hat{\mu}_{dk})^{T}\hat{\Sigma}^{-1}_{dk}(\Delta_{d}-\hat{\mu}_{dk})\right)}{\sqrt{2\pi\hat{\Sigma}_{dk}}}.
\end{aligned}
\end{equation}

In our experiments, we assume $\hat{\Sigma}_{dk}$ are diagonal matrices and estimate the GMM parameters using EM algorithm~\cite{dempster1977maximum}. We enforce the learning objective on the top layer of our graph convolutional network and learn the model parameters $W$ with a back-propagation technique.

\subsection{3D Pose estimation}
We optimize the 3D pose estimation, where the 3D pose is derived from the output mesh. Assume we have an output mesh, which is computed from the graph convolutional neural network. We regress the output mesh to the 3D pose, and minimize the error between the predicted 3D pose $J_{3D}$ and the ground truth $\bar{J}_{3D}$. Similar to previous study~\cite{kolotouros2019convolutional}, we apply L1 loss function to achieve this objective:
\begin{equation}%
\begin{aligned}%
\label{eqn:mse-loss}%
\mathcal{L}_{3DPose}(W) = \frac{1}{K}\sum_{i=1}^{K} \left| \left| J_{3D}-\bar{J}_{3D} \right| \right|_1,
\end{aligned}
\end{equation}
where $K$ is the total number of joints.

\subsection{2D Pose estimation}

In addition to 3D pose estimation, we enhance the performance of pose estimation by projecting the 3D pose to the 2D pose using the weak-perspective projection with the predicted camera parameters. Following the previous works~\cite{kolotouros2019convolutional,kanazawa2018end}, the camera parameters consist of a scaling factor and a 2D translation. The camera parameters are regressed using the graph convolutional neural network. We then minimize the prediction error between the predicted 2D pose $J_{2D}$ and the ground truth $\bar{J}_{2D}$. 

\begin{equation}%
\begin{aligned}%
\label{eqn:mse-loss}%
\mathcal{L}_{2DPose}(W) = \frac{1}{K}\sum_{i=1}^{K} \left| \left| J_{2D}-\bar{J}_{2D} \right| \right|_1.
\end{aligned}
\end{equation}

\subsection{2D Part segmentation}

Inspired by previous studies~\cite{furukawa2009accurate,seitz2006comparison,rivers20103d,kato2018renderer} that have shown the effectiveness of using silhouette information for 3D object modeling, we add the part segmentation in our loss function. 
Given the predicted camera parameters, we project the output mesh to the 2D part segmentation masks $B_{2D}$, and minimize the difference between the predicted part segmentation masks $B_{2D}$ and the ground truth masks $\bar{B}_{2D}$. We apply Mean Square Error (MSE) loss function to obtain the objective:\begin{equation}\begin{aligned}%
\label{eqn:mse-loss}%
\mathcal{L}_{2DPart}(W) = \frac{1}{Z}\sum_{i=1}^{Z} \left| \left| B_{2D}-\bar{B}_{2D} \right| \right|_2^2,
\end{aligned}
\end{equation}
where $Z$ is the number of body part categories. To achieve end-to-end training, we use a differentiable rendering model~\cite{kato2018renderer} to render the part segmentation masks, and approximate the gradients for back propagation. 


\subsection{Model architecture}

Figure~\ref{fig:first} illustrates the proposed model. Our model takes an image of size $224\times224$ as input, and predicts a set of mesh vertices $Y$. The model consists of three subnetworks: \textit{Pose-Part Network}, \textit{Feature Embedding Network} and \textit{Graph CNN}. 

{\flushleft \textbf{Pose-Part Network.}} We use a network similar to the existing multi-task networks~\cite{he2017mask,lin2019cross} to predict the 2D pose and part segmentation. 
We denote $\mathcal{M}$ as our Pose-Part Network, and its outputs are $\{H_{2D}, B_{2D}\} = \mathcal{M}(I)$, where $I$ is an input image, $H_{2D}$ denotes the pose estimation heatmaps, and $B_{2D}$ denotes the part segmentation masks.

{\flushleft \textbf{Feature Embedding network.}} 
The inputs to the Feature Embedding Network include the input image $I$, the pose heatmaps $H_{2D}$, and the part segmentation masks $B_{2D}$. It outputs feature vector $X$ as the input of the Graph CNN.
In this work, we use a ResNet50 network~\cite{he2016deep}, and extract a 2048-dimension feature vector. We denote $\mathcal{E}$ as our Feature Embedding Network, and its output is $X = \mathcal{E}(I, H_{2D}, B_{2D})$.


{\flushleft \textbf{Graph CNN.}} We estimate the 3D coordinates of the mesh vertices by using the graph convolutional neural network (Graph CNN). Our Graph CNN is in spirit similar to~\cite{kolotouros2019convolutional}, and we do not have a SMPL regression network.
Given the feature vector $X$ extracted from our Feature Embedding Network, we attach $X$ to the 3D coordinates of each vertex in the graph. Then, we perform a series of convolutions on the graph and output the mesh vertices $Y$ and the weak-perspective camera parameters $c_w=[s,t_x,t_y]$, where $s$, $t_x$, $t_y$ indicate the scaling factor and translation of two directions, respectively. We denote $\mathcal{R}$ as our Graph CNN, and the outputs of our Graph CNN are $\{Y,c_w\} = \mathcal{R}(X)$.



\subsection{Training}


We train our model in an end-to-end fashion, and update the model parameters using a back-propagation technique. We apply the proposed loss functions on the output of the Graph CNN $\mathcal{R}$. We also apply intermediate supervision on the Pose-Part Network $\mathcal{M}$ for learning pose estimation and part segmentation. We use an Adam optimizer with a learning rate $3\times10^{-4}$, and the batch size is $32$. In the experiments, we first pre-train our Pose-Part Network using MSCOCO~\cite{lin2014microsoft} and Pascal-Person-Parts~\cite{chen2014detect} to ensure a reasonable performance for pose estimation and part segmentation. Next, we train our full model using UP-3D~\cite{lassner2017unite} and Human3.6M~\cite{ionescu2014human3} datasets to learn mesh reconstruction. Although some of the existing datasets have the 3D mesh annotations, we do not use the ground truth meshes for training. \textcolor{black}{To have a fair performance comparison, we follow the previous studies~\cite{lassner2017unite,omran2018neural,kolotouros2019convolutional,Rong_2019_ICCV,kanazawa2018end,pavlakos2018learning} and use the same topology as the SMPL model~\cite{loper2015smpl} in the experiments. It is worth noting that our method does not have restrictions on the mesh topology, and can be extended to other human mesh that does not have SMPL parameters.}



\begin{figure*}
\setlength{\tabcolsep}{1pt}
\centering
\begin{tabular}{cccccc}
		\includegraphics[width=0.33\columnwidth]{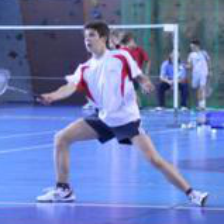} &
	\includegraphics[width=0.33\columnwidth]{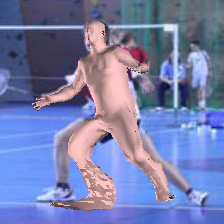} &
	\includegraphics[width=0.33\columnwidth]{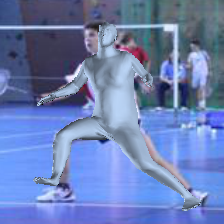} &
	\includegraphics[width=0.33\columnwidth]{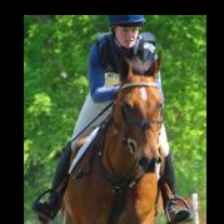} &
	\includegraphics[width=0.33\columnwidth]{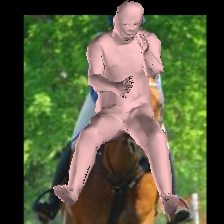} &
	\includegraphics[width=0.33\columnwidth]{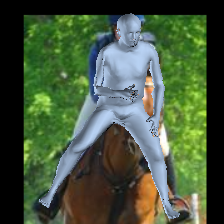}
		 \\
	
	\includegraphics[width=0.33\columnwidth]{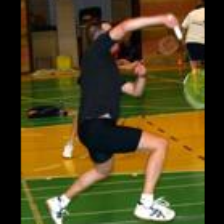} &
	\includegraphics[width=0.33\columnwidth]{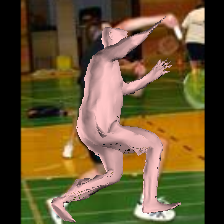} &
	\includegraphics[width=0.33\columnwidth]{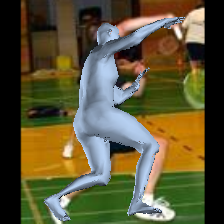} &
		\includegraphics[width=0.33\columnwidth]{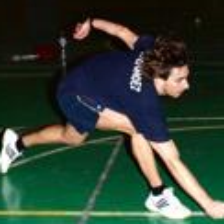} &
	\includegraphics[width=0.33\columnwidth]{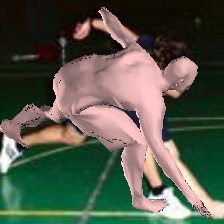} &
	\includegraphics[width=0.33\columnwidth]{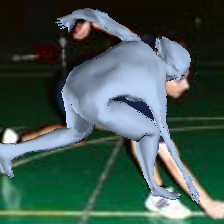} \\
	
	\includegraphics[width=0.33\columnwidth]{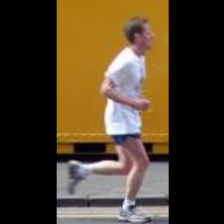} &
	\includegraphics[width=0.33\columnwidth]{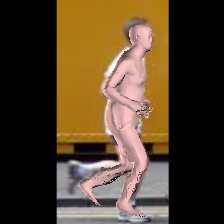} &
	\includegraphics[width=0.33\columnwidth]{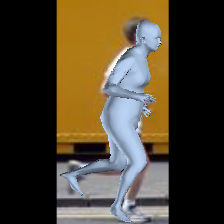} &

	\includegraphics[width=0.33\columnwidth]{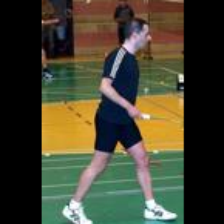} &
	\includegraphics[width=0.33\columnwidth]{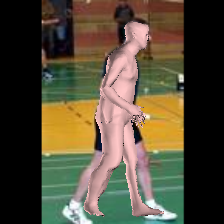} &
	\includegraphics[width=0.33\columnwidth]{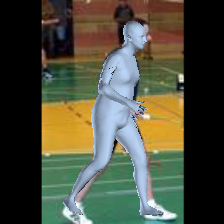}\\

	Input & GraphCMR & Ours & Input & GraphCMR & Ours
\end{tabular}
	\captionof{figure}{
	Qualitative comparison with the state-of-the-art nonparametric approach on the UP-3D dataset. Light blue color indicates the results of the proposed method, and light pink color indicates the results of GraphCMR~\cite{kolotouros2019convolutional}. Without using ground truth meshes in the training, our method achieves comparable or better performance than the state-of-the-art method which requires ground truth meshes.
	}
	\label{fig:qual}
\end{figure*}

\section{Experimental Results}\label{sec:exp}

\subsection{Evaluation benchmarks}
 
\textbf{UP-3D~\cite{lassner2017unite}} is an outdoor-image dataset with rich annotations including 3D pose, 2D pose, part segmentation, and mesh ground truthes. The images are collected from
2D human pose benchmarks, such as MPII~\cite{andriluka14cvpr} and LSP~\cite{Johnson10}. The annotations are created by performing shape fitting on each human in the image. We train our model using UP-3D training data, and evaluate the performance using the metric of mean Per-Vertex-Error (mPVE) on the UP-3D test set.

\textbf{Human3.6M~\cite{ionescu2014human3}} is an indoor large-scale dataset with 3D pose annotations. Each image has a subject performing a different action. Following the common setting~\cite{kolotouros2019convolutional}, we use the subjects S1, S5, S6, S7 and S8 for training, and use the subjects S9 and S11 for testing. 

\textbf{LSP~\cite{Johnson10}} is an outdoor-image dataset. We evaluate part segmentation performance on LSP test set, where the segmentation labels are provided by Lassner~\textit{et al.}\cite{lassner2017unite}.

\textcolor{black}{\textbf{3DPW~\cite{vonMarcard2018}} is an outdoor large-scale dataset with mesh ground truthes. We evaluate the robustness of our method with cross-dataset evaluation, i.e., trained on UP-3D dataset and applied to 3DPW dataset.}

\subsection{Main results}
\begin{table}
\centering
\begin{tabular}{lc}
    \toprule
	Method  & mean Per-Vertex-Error\\
	\midrule
	Lassner~\textit{et al.}\cite{lassner2017unite} & $169.8$\\ NBF~\cite{omran2018neural}& $134.6$  \\
	HMR~\cite{kanazawa2018end} & $149.2$  \\
	DC~\cite{Rong_2019_ICCV} & $137.5$ \\
	Pavlakos~\textit{et al.}\cite{pavlakos2018learning} & $100.5$\\
	GraphCMR~\cite{kolotouros2019convolutional} & $100.2$ \\
	\midrule
    Ours & $81.5$\\
	Ours + GT Inputs & $\textbf{73.7}$\\
	\bottomrule
\end{tabular}
\caption{\textcolor{black}{Performance comparison of human mesh reconstruction using metric mean Per-Vertex-Error (mPVE) on UP-3D test set. The unit is millimetter (mm).}}
\label{tbl:compare}
\end{table}

We compare the performance of our method with the state-of-the-art approaches which require either ground truth meshes or 2D-to-3D dense correspondence labels, and Table~\ref{tbl:compare} shows the performance comparison on UP-3D dataset. We evaluate the performance of mesh reconstruction by using the metric mean Per-Vertex-Error (mPVE)~\cite{pavlakos2018learning}, where the unit is millimeter (mm). \textcolor{black}{For each mesh vertex, we estimate the Euclidean distance between the ground-truth location and the predicted location. We average over all the vertices to provide a mean Per-Vertex-Error (mPVE). Our method outperforms the previous state-of-the-art approaches by a significant margin.}

\begin{table*}[h]
\centering
\resizebox{1.\textwidth}{!}{
\begin{tabular}{lcccccc}
    \toprule
	Method & DTCrossStreets & DTRampAndStairs & DTRunForBus & DTWarmWelcome & OutdoorsFencing & CourtyardDancing\\
	\midrule
	GraphCMR~\cite{kolotouros2019convolutional} & 85.22 & 86.69 & 70.57 & 85.02 & 73.90 & 112.36 \\
	Ours & \textbf{83.47} & \textbf{83.99} & \textbf{68.87} & \textbf{83.92} & \textbf{70.88} & \textbf{71.04}\\
	\bottomrule
\end{tabular}
}
\captionof{table}{\textcolor{black}{Performance comparison of human mesh reconstruction using metric mean Per-Vertex-Error (mPVE) on 3DPW sequences. The unit is millimeter (mm).}}
\label{tbl:3dpw}
\end{table*}

If we use the ground truth body priors (ground truth labels for both pose estimation heatmaps and part segmentation masks) as the inputs of our Feature Embedding Network, we obtain an additional gain and the result is shown in the bottom row of Table~\ref{tbl:compare}. This is an indication that pose estimation and part segmentation are useful for human mesh reconstruction.

Figure~\ref{fig:qual} shows the qualitative comparisons with the state-of-the-art nonparametric approach (GraphCMR~\cite{kolotouros2019convolutional}) which also uses graph convolutional neural network but
directly regresses the ground truth mesh vertices. The results show that, without using the ground truth meshes, our method is on par or even slightly better than the existing techniques.

\textcolor{black}{We evaluate the robustness of the proposed method on the 3DPW dataset~\cite{vonMarcard2018}. Table~\ref{tbl:3dpw} shows the performance comparison with the state-of-the-art nonparametric approach~\cite{kolotouros2019convolutional}. Both models are trained using UP-3D and Human3.6M but without 3DPW dataset. Our method does not use any of the 3D ground truth meshes in either UP-3D or Human3.6M, while GraphCMR~\cite{kolotouros2019convolutional} used the ground truth meshes of both UP-3D and Human3.6M. For a fair comparison, we use ground truth bounding boxes to crop the persons as the inputs for the two methods. Our method performs comparably or better than GraphCMR~\cite{kolotouros2019convolutional}.}

We evaluate the 3D pose of the reconstructed mesh by comparing the performance of 3D pose estimation on Human3.6M dataset~\cite{ionescu2014human3} using Protocol 2 Reconstruction Error metric~\cite{ionescu2014human3,zhou2018monocap,martinez2017simple}, where the unit is millimeter (mm). In Table~\ref{tbl:h36}, the upper-rows show the state-of-the-art results that try to regress SMPL parameters for human mesh reconstruction. The bottom two rows show the comparison of our method with the state-of-the-art nonparametric method that does not regress SMPL parameters. Our method does not use any of the ground truth meshes in training, and achieves comparable or even better performance than several baseline approaches that require Human3.6M ground truth meshes. 

\begin{table}
\centering
\begin{tabular}{lcc}
    \toprule
	Method  & SMPL & Reconst. Error (mm)\\
	\midrule
	Lassner~\textit{et al.}\cite{lassner2017unite} & \cmark & $93.9$\\
	SMPLify~\cite{bogo2016keep} & \cmark &$82.3$\\
	Pavlakos~\textit{et al.}~\cite{pavlakos2018learning} & \cmark & $75.9$\\
	HMR unpaired~\cite{kanazawa2018end} & \cmark & $66.5$\\
	NBF~\cite{omran2018neural}& \cmark & $59.9$  \\
	HMR~\cite{kanazawa2018end} & \cmark & $56.8$\\
	GraphCMR+SMPL~\cite{kolotouros2019convolutional} & \cmark & $50.1$\\
	\midrule
	GraphCMR~\cite{kolotouros2019convolutional} & \xmark & $69.0$\\
	Ours & \xmark & $58.5$\\
	\bottomrule
\end{tabular}
\caption{Evaluation of 3D pose estimation on Human3.6M dataset using Protocol 2. The results are Reconstruction errors in millimeter (mm). Our approach is competitive with the state-of-the-art approaches.}
\label{tbl:h36}
\end{table}

\begin{table}
\centering
\begin{tabular}{lcccc}
    \toprule
    & \multicolumn{2}{c}{FB Seg.} & \multicolumn{2}{c}{Part Seg.}\\
	Method  & Accuracy & F1 & Accuracy & F1\\
	\midrule
	SMPLify~\cite{bogo2016keep} & $91.89$ & $0.88$ & $87.71$ & $0.64$\\
	SMPLify on~\cite{pavlakos2018learning} & $92.17$ & $0.88$ & $88.24$ & $0.64$\\
	BodyNet~\cite{varol2018bodynet} & $92.75$ & $0.84$ & $-$ & $-$\\
	\midrule
	HMR~\cite{kanazawa2018end} & $91.67$ & $0.87$ & $87.12$ & $0.60$\\
	GraphCMR~\cite{kolotouros2019convolutional} & $91.46$ & $0.87$ & $88.69$ & $0.66$\\
	Ours & $91.23$ & $0.86$ & $88.86$ & $0.66$\\
	\bottomrule
\end{tabular}
\caption{Performance comparison of segmentation on LSP test set. The numbers are accuracy scores and F1 scores. The top three rows show the approaches that perform some optimization (post)-processing. The bottom three rows show the comparison with the regression-based approaches. Without using ground truth meshes in training, our approach is competitive with the state-of-the-art methods.}
\label{tbl:lsp}
\end{table}

We also evaluate the 3D shape by comparing the performance of part segmentation on LSP test set. Following the common settings~\cite{kanazawa2018end,kolotouros2019convolutional}, we report the segmentation accuracy and the average F1 score for $6$ body parts and the background in Table~\ref{tbl:lsp}. We also report the results on foreground-background segmentation. Our method achieves comparable or better performance than the state-of-the-arts approaches that use ground truth meshes in training.

\begin{figure*}[h]
\centering
\setlength{\tabcolsep}{1pt}
\begin{tabular}{cccccc}
	
	\includegraphics[width=0.16\textwidth]{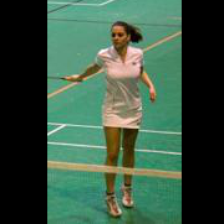} &
	\includegraphics[width=0.16\textwidth]{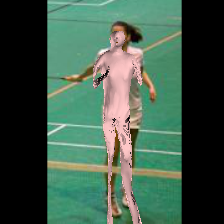} &
	\includegraphics[width=0.16\textwidth]{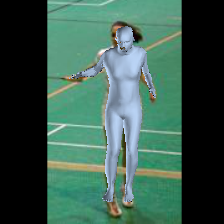} &
	\includegraphics[width=0.16\textwidth]{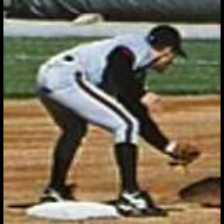} &
	\includegraphics[width=0.16\textwidth]{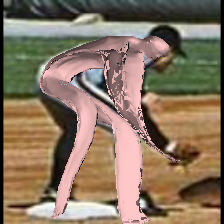} &
	\includegraphics[width=0.16\textwidth]{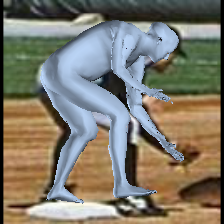}\\
	 &
	\includegraphics[width=0.16\textwidth]{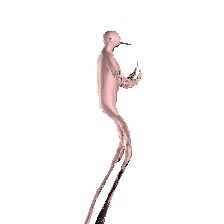} &
	\includegraphics[width=0.16\textwidth]{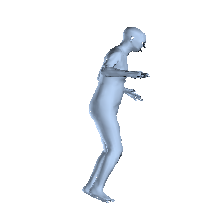} &
	&
	\includegraphics[width=0.16\textwidth]{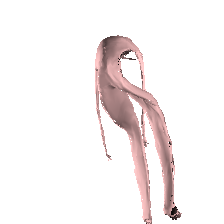} &
	\includegraphics[width=0.16\textwidth]{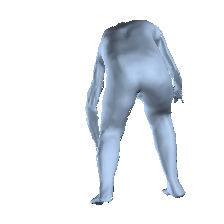}\\
	\textit{\small{Input}} & \textit{\small{w/o Laplacian}} & \textit{\small{w/ Laplacian}} & \textit{\small{Input}} & \textit{\small{w/o Laplacian}} & \textit{\small{w/ Laplacian}}
\end{tabular}
	\captionof{figure}{
	Qualitative comparison of our method using different training configurations.}
	\label{fig:laplaican-abl}
\end{figure*}

\begin{table}
\centering
\begin{tabular}{lc}
    \toprule
	Method  & mean Per-Vertex-Error\\
	\midrule
	\textit{Ours w/o Laplacian prior}& $240.3$\\
    \textit{Ours} & $81.5$\\
	\bottomrule
\end{tabular}
\caption{Ablation study of proposed Laplacian prior, evaluated on UP-3D test set with mean Per-Vertex-Error. The unit is millimeter (mm).}
\label{tbl:abl}
\end{table}
\begin{table}
\centering
\resizebox{\columnwidth}{!}{
\begin{tabular}{ccc}
    \toprule
	Laplacian prior  & Part Seg Loss & mean Per-Vertex-Error\\
	\midrule
	\xmark  & \cmark & $240.3$\\
	\cmark  & \xmark & $91.3$\\
	\cmark  & \cmark & $81.5$\\
	\bottomrule
\end{tabular}
}
\caption{\textcolor{black}{Ablation study of the proposed two loss terms, evaluated on UP-3D test set with mean Per-Vertex-Error. The unit is millimeter (mm).}}
\label{tbl:abalation}
\end{table} 

\subsection{Ablation study}
{\flushleft \textbf{Laplacian prior.}} Since our approach learns with the Laplacian prior, one interesting question is whether the proposed learning objective is useful. To answer this question, we have trained our network without the Laplacian prior (i.e. with pose and segmentation losses only). This configura-
tion is denoted as \textit{w/o Laplacian}, and the results on UP-3D are shown in Table~\ref{tbl:abl}. We can see that Laplacian prior loss is critical to our learning objective for human mesh reconstruction. Figure~\ref{fig:laplaican-abl} shows a qualitative comparison of the two configurations. It can be seen that training without Laplacian prior term produces wrong body shape.

\textcolor{black}{{\flushleft \textbf{Part segmentation loss.}} We also evaluate the effectiveness of the proposed part segmentation loss, and Table~\ref{tbl:abalation} shows the comparison. For completeness, we also show the results of training with the Laplacian prior. We can see that training only with part segmentation loss does not work well, and Laplacian prior further improves the results. Our model achieves the best performance when two proposed loss terms are used.}

{\flushleft \textbf{Pose-Part Network.}} Since our Pose-Part Network predicts pose heatmaps and part segmentation masks, one may wonder whether this is useful. To answer the question, we train our model without Pose-Part Network, and Table~\ref{tbl:abl2} shows the results. We can see that pose heatmaps and part segmentation masks significantly improve the learning.

\begin{figure}
\centering
\setlength{\tabcolsep}{1pt}
\begin{tabular}{cccc}
    & \footnotesize{Input Image} & \footnotesize{Output mesh} & \footnotesize{Output mesh in another view}\\
	\rotatebox[origin=l]{90}{\small{Lap. Smoothness}} &
	\includegraphics[width=0.29\columnwidth]{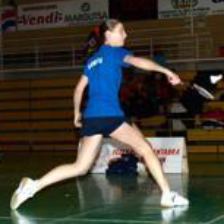} &
	\includegraphics[width=0.29\columnwidth]{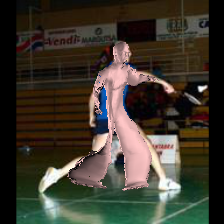} &
	\includegraphics[width=0.29\columnwidth]{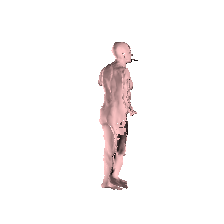}\\
	\rotatebox[origin=l]{90}{\small{Laplacian Prior}} &
	\includegraphics[width=0.29\columnwidth]{./smoothness/8460_00923_imagepng_input.png} &
	\includegraphics[width=0.29\columnwidth]{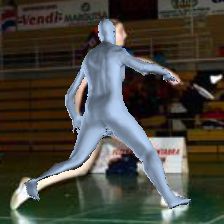} &
	\includegraphics[width=0.29\columnwidth]{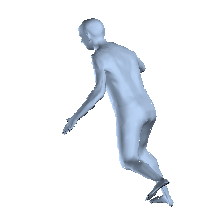}\\
\end{tabular}
	\captionof{figure}{
	Comparison of Laplacian smoothness and the proposed Laplacian prior.}
	\label{fig:laplaican-smooth}
\end{figure}


\textcolor{black}{{\flushleft \textbf{Analysis of different regularizers.}} Laplacian smoothness is commonly used in the literature~\cite{wang2018pixel2mesh,kanazawa2018learning} as a regularizer to avoid self-intersections for 3D object modeling. One may wonder what if we replace the proposed Laplacian prior with the Laplacian smoothness. We have conducted this experiment and Figure~\ref{fig:laplaican-smooth} shows the qualitative comparison. We can see that training with Laplacian smoothness produces wrong results. This is because Laplacian smoothness term is not a strong enough regularizer, and as a result the training process usually gets stuck in a local minimum.}
\begin{figure*}[t]
\includegraphics[width=0.140\columnwidth]{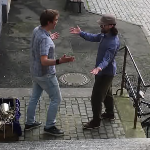}
\includegraphics[width=0.140\columnwidth]{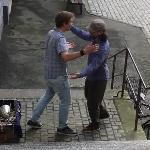}
\includegraphics[width=0.140\columnwidth]{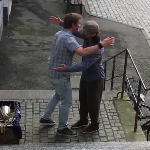}
\includegraphics[width=0.140\columnwidth]{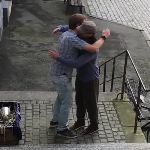}
\includegraphics[width=0.140\columnwidth]{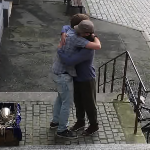}
\includegraphics[width=0.140\columnwidth]{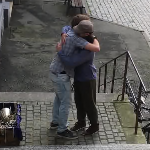}
\includegraphics[width=0.140\columnwidth]{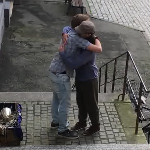}
\includegraphics[width=0.140\columnwidth]{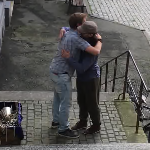}
\includegraphics[width=0.140\columnwidth]{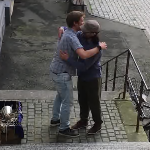}
\includegraphics[width=0.140\columnwidth]{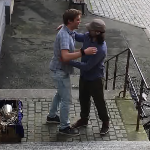}
\includegraphics[width=0.140\columnwidth]{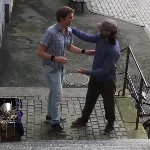}
\includegraphics[width=0.140\columnwidth]{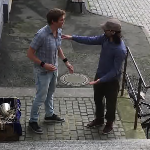}
\includegraphics[width=0.140\columnwidth]{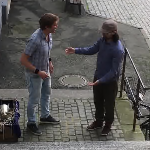}
\includegraphics[width=0.140\columnwidth]{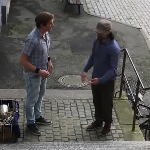}\\
\includegraphics[width=0.140\columnwidth]{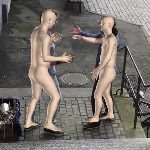}
\includegraphics[width=0.140\columnwidth]{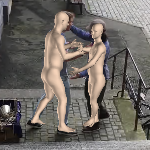}
\includegraphics[width=0.140\columnwidth]{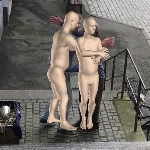}
\includegraphics[width=0.140\columnwidth]{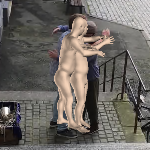}
\includegraphics[width=0.140\columnwidth]{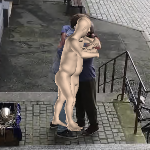}
\includegraphics[width=0.140\columnwidth]{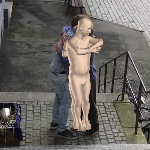}
\includegraphics[width=0.140\columnwidth]{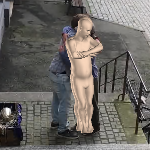}
\includegraphics[width=0.140\columnwidth]{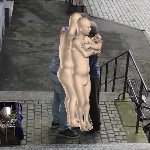}
\includegraphics[width=0.140\columnwidth]{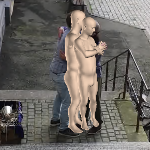}
\includegraphics[width=0.140\columnwidth]{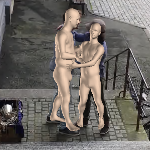}
\includegraphics[width=0.140\columnwidth]{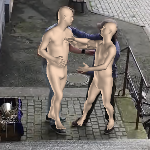}
\includegraphics[width=0.140\columnwidth]{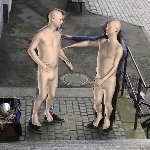}
\includegraphics[width=0.140\columnwidth]{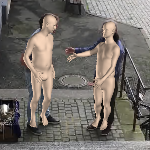}
\includegraphics[width=0.140\columnwidth]{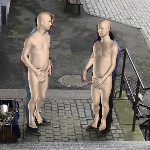}\\
\includegraphics[width=0.140\columnwidth]{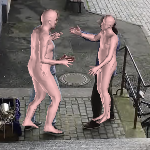}
\includegraphics[width=0.140\columnwidth]{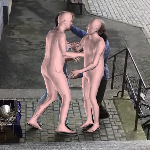}
\includegraphics[width=0.140\columnwidth]{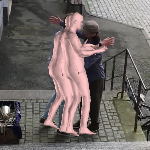}
\includegraphics[width=0.140\columnwidth]{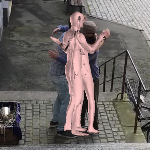}
\includegraphics[width=0.140\columnwidth]{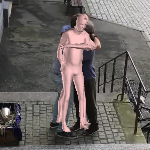}
\includegraphics[width=0.140\columnwidth]{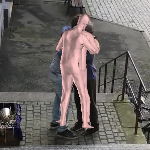}
\includegraphics[width=0.140\columnwidth]{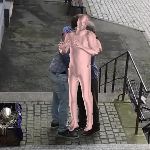}
\includegraphics[width=0.140\columnwidth]{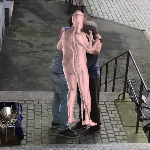}
\includegraphics[width=0.140\columnwidth]{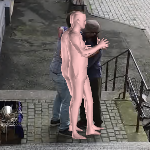}
\includegraphics[width=0.140\columnwidth]{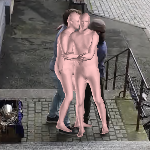}
\includegraphics[width=0.140\columnwidth]{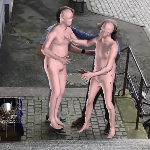}
\includegraphics[width=0.140\columnwidth]{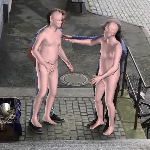}
\includegraphics[width=0.140\columnwidth]{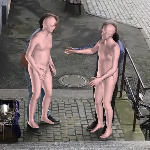}
\includegraphics[width=0.140\columnwidth]{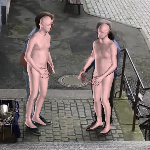}\\
\includegraphics[width=0.140\columnwidth]{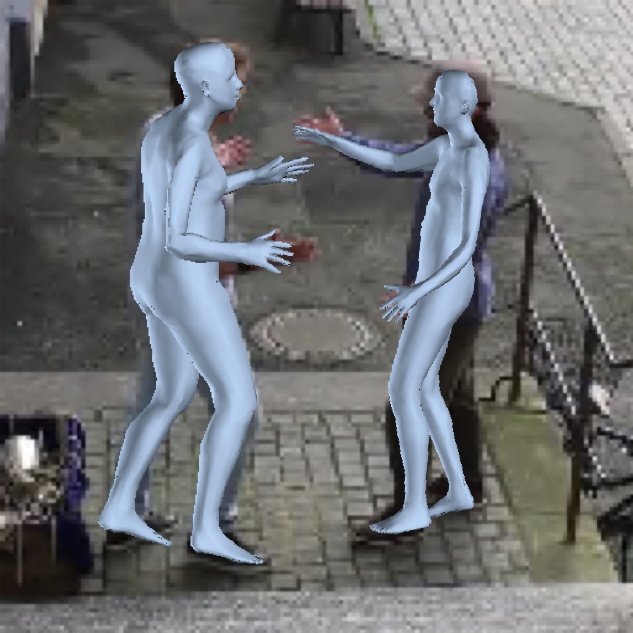}
\includegraphics[width=0.140\columnwidth]{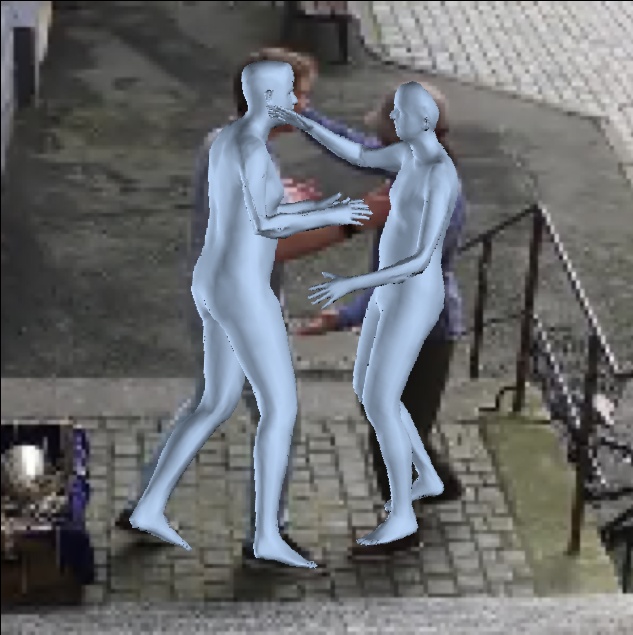}
\includegraphics[width=0.140\columnwidth]{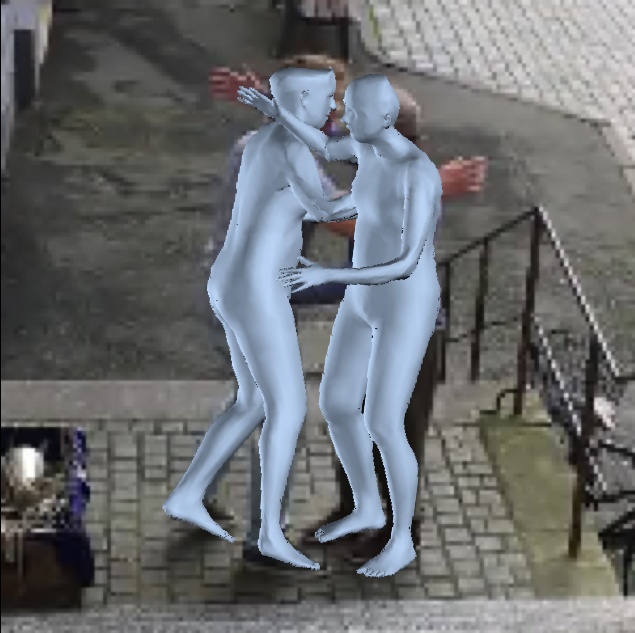}
\includegraphics[width=0.140\columnwidth]{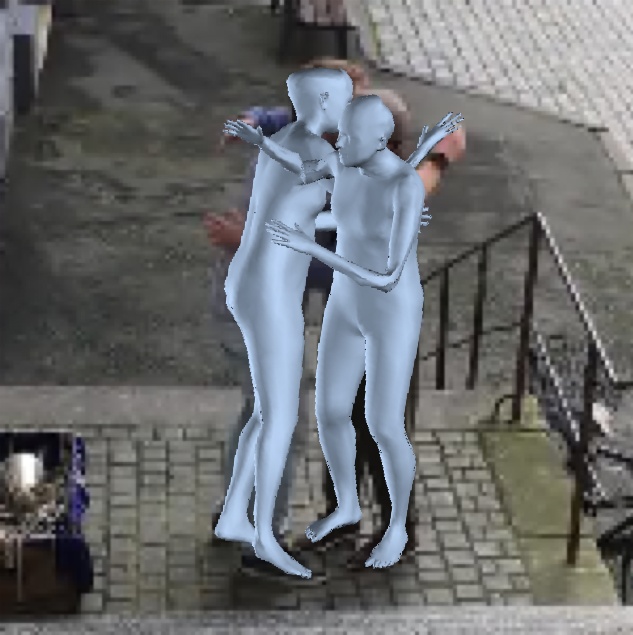}
\includegraphics[width=0.140\columnwidth]{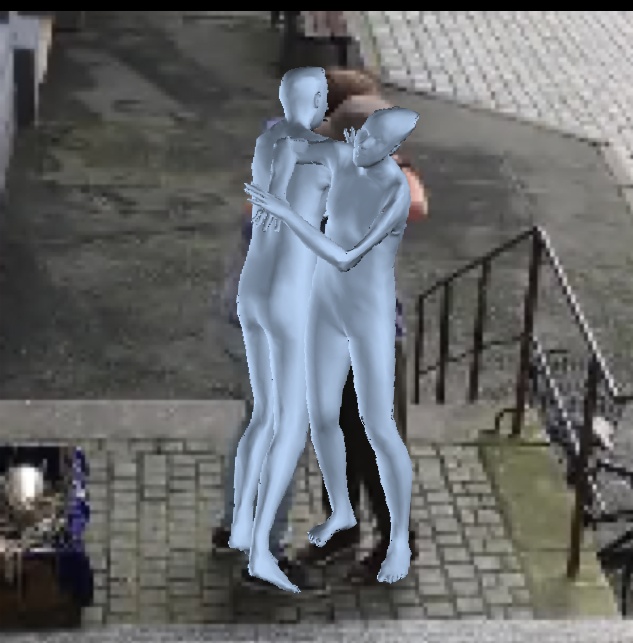}
\includegraphics[width=0.140\columnwidth]{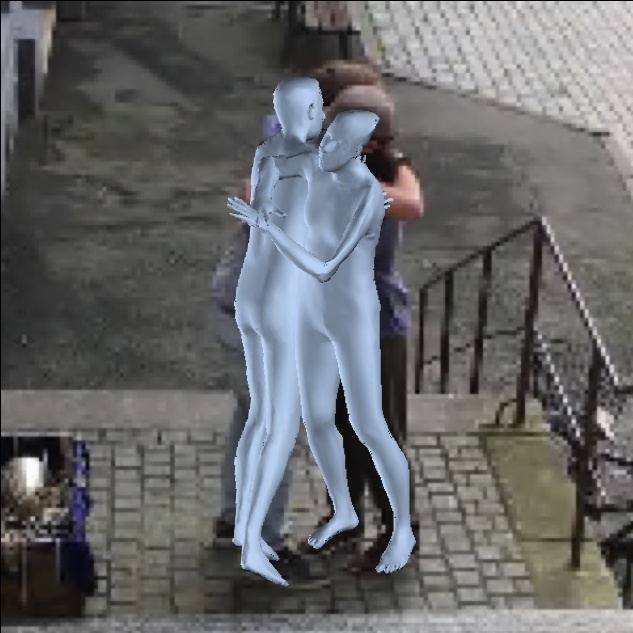}
\includegraphics[width=0.140\columnwidth]{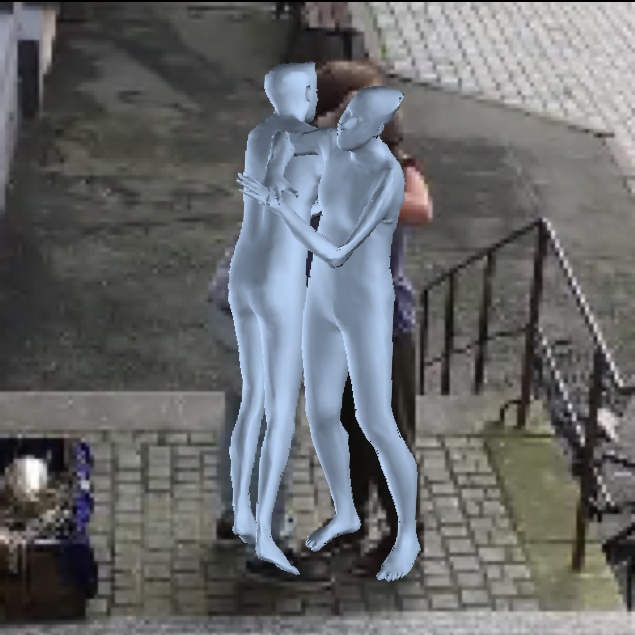}
\includegraphics[width=0.140\columnwidth]{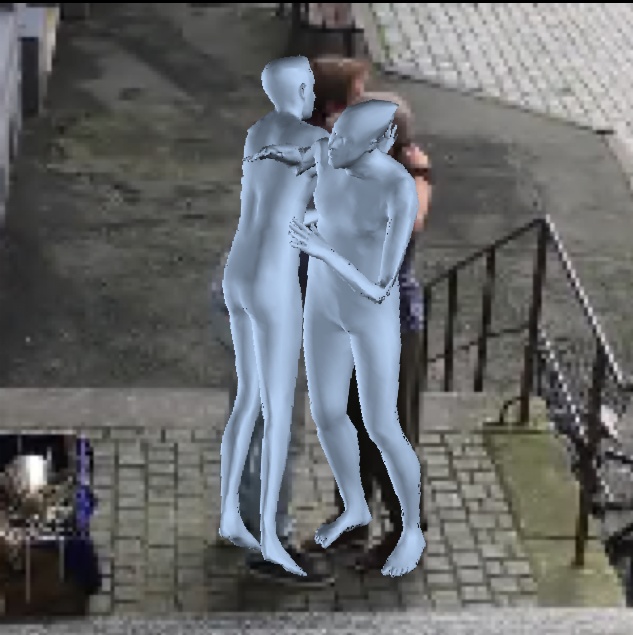}
\includegraphics[width=0.140\columnwidth]{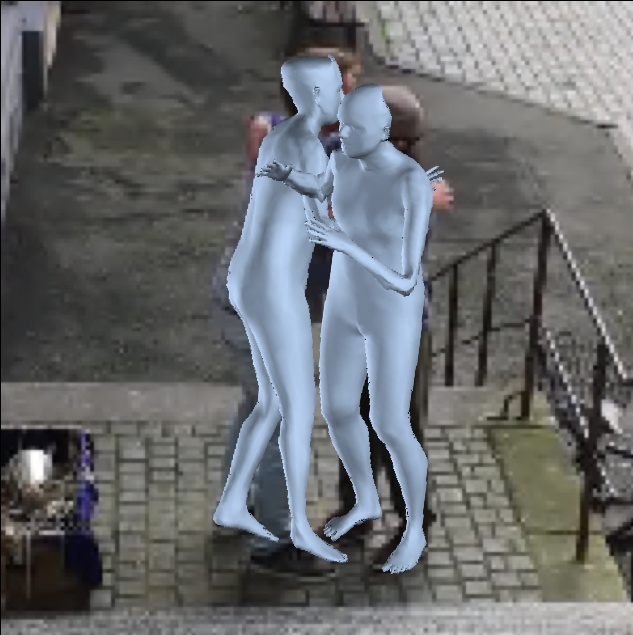}
\includegraphics[width=0.140\columnwidth]{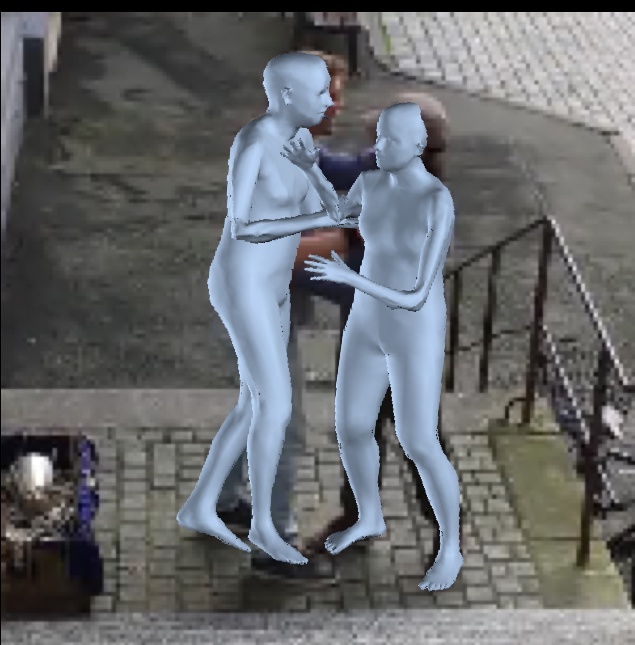}
\includegraphics[width=0.140\columnwidth]{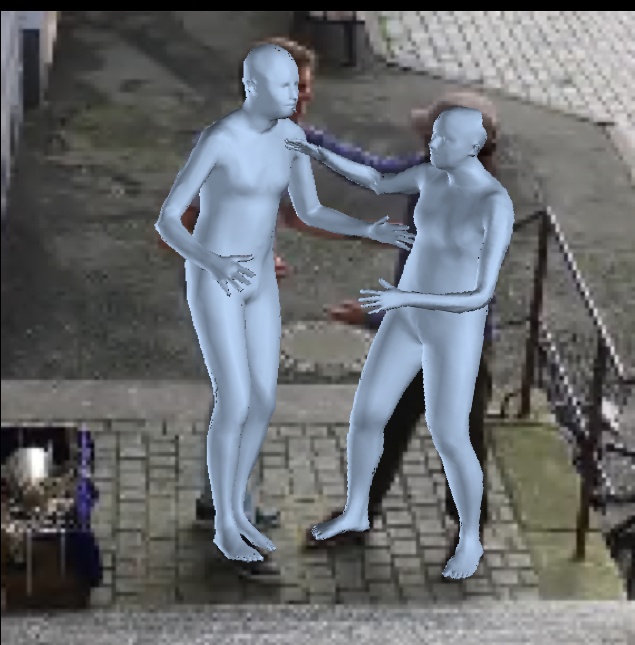}
\includegraphics[width=0.140\columnwidth]{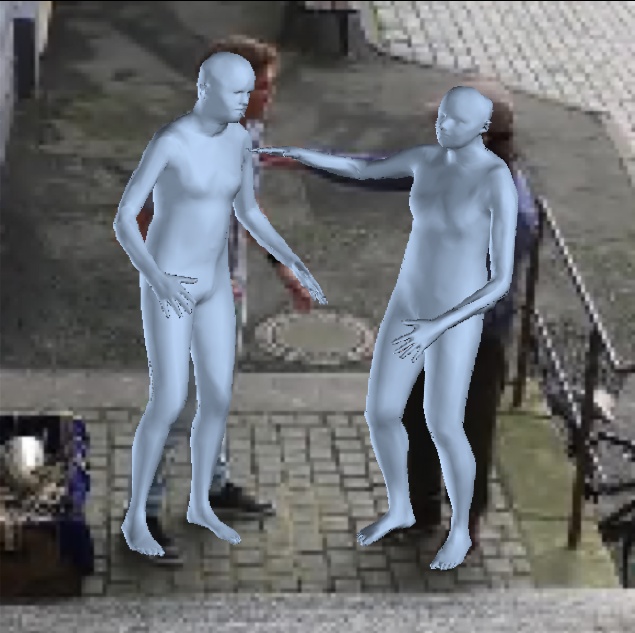}
\includegraphics[width=0.140\columnwidth]{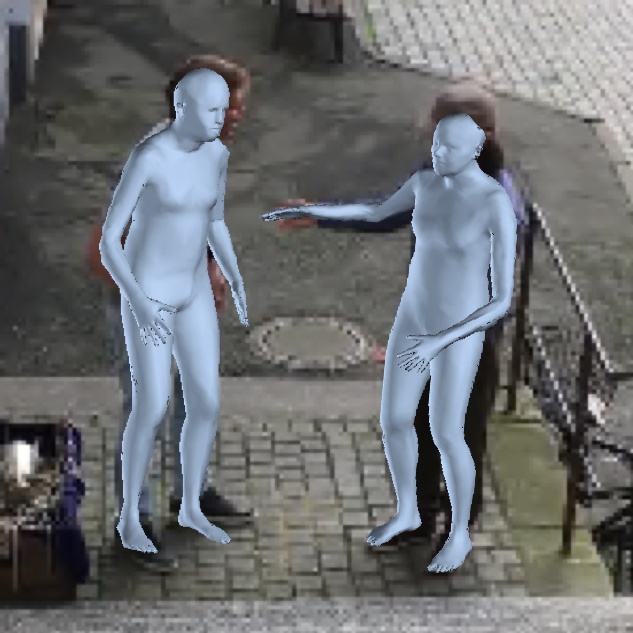}
\includegraphics[width=0.140\columnwidth]{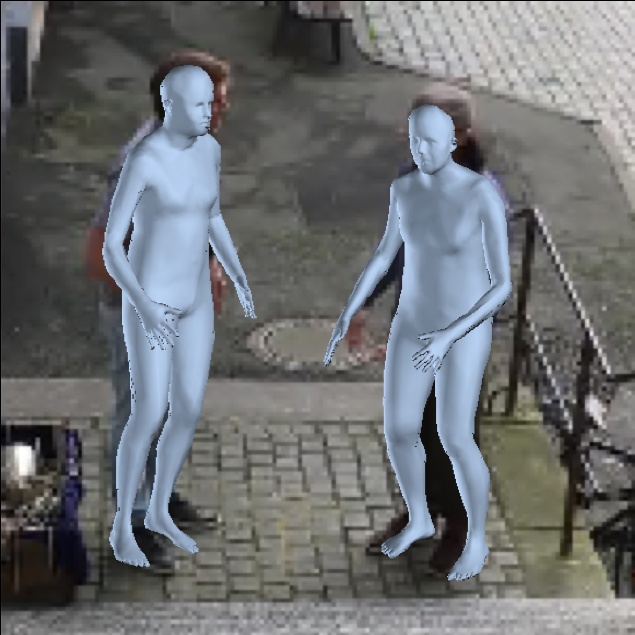}
	\captionof{figure}{
	Qualitative comparison with the previous state-of-the-art approaches~\cite{kanazawa2018end,kolotouros2019convolutional} on the challenging 3DPW dataset~\cite{vonMarcard2018}. \textcolor{black}{The top row shows two people embracing each other. The second row shows the results of a representative parametric approach HMR~\cite{kanazawa2018end}. The third row shows the results of the previous state-of-the-art nonparametric approach GraphCMR~\cite{kolotouros2019convolutional}.} The bottom row shows our results. Previous approaches failed to reconstruct the mesh for the two persons due to occlusions. In contrast, our method reconstructs correct human meshes for both people in all the frames.}
	\label{fig:compare-parametric}
\end{figure*}
\begin{table}
\centering
\begin{tabular}{lc}
    \toprule
	Method  & mean Per-Vertex-Error\\
	\midrule
	\textit{Ours w/o Pose-Part Net}& $110.0$\\
    \textit{Ours}& $81.5$\\
	\bottomrule
\end{tabular}
\caption{Ablation study of the Pose-Part Network, also evaluated on UP-3D test set with mean Per-Vertex-Error. The unit is millimeter (mm).}
\label{tbl:abl2}
\end{table}
\begin{table}
\centering
\begin{tabular}{lc}
    \toprule
	Method & mean Per-Vertex-Error\\
	\midrule
	Ours, without GT meshes & 81.5\\
	Ours, with GT meshes & 65.1\\
	\bottomrule
\end{tabular}
\caption{\textcolor{black}{Ablation study of our method with and without using GT meshes in training, evaluated on UP-3D test set with mean Per-Vertex-Error. The unit is millimeter (mm).}}
\label{tbl:abalation3}
\end{table}
\textcolor{black}{{\flushleft \textbf{Extension to supervised training.}} We study the \textit{upper bound} performance of our method when ground truth meshes are available for training, and Table~\ref{tbl:abalation3} shows the results. We add a vertex regression loss to our learning objective (Eq.(\ref{eqn:overall-loss})), and train our model with the ground truth meshes provided in UP-3D training set. Our model improves the previous state-of-the-art performance to $65$ mPVE on UP-3D test set.}

\subsection{Qualitative comparison}

We conduct qualitative comparisons with the state-of-the-art methods~\cite{kolotouros2019convolutional,kanazawa2018end} on the challenging 3DPW dataset~\cite{vonMarcard2018}, and Figure~\ref{fig:compare-parametric} shows the results. We can see that previous state-of-the-art approaches~\cite{kolotouros2019convolutional,kanazawa2018end} had difficulties to reconstruct the mesh of the person on the right due to occlusions. \textcolor{black}{They reconstructed a mesh for the person on the right but is not correct, and they failed completely from the fourth column to the eight column where the occlusions are more severe.} Our method reconstructs correct human meshes for both people even though there are quite severe occlusions between them. By explicitly feeding 2D pose heatmaps and part segmentation masks into the Graph CNN feature embedding, the robustness of our method to occlusions has been significantly improved.

\begin{figure*}
 \centering
 \setlength{\tabcolsep}{0.5pt}
 \begin{tabular}{cccc}
  	\includegraphics[width=.25\textwidth]{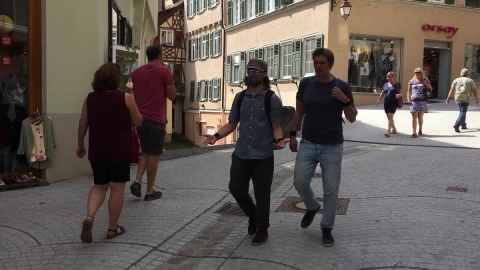}
    &\includegraphics[width=.25\textwidth]{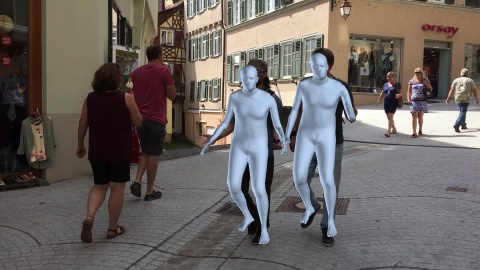}
  	&\includegraphics[width=.25\textwidth]{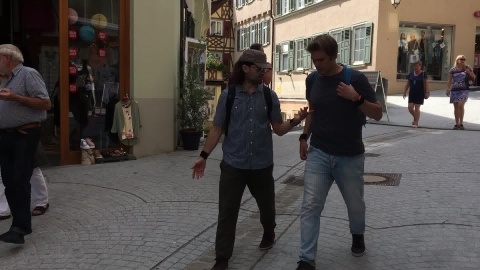}
    &\includegraphics[width=.25\textwidth]{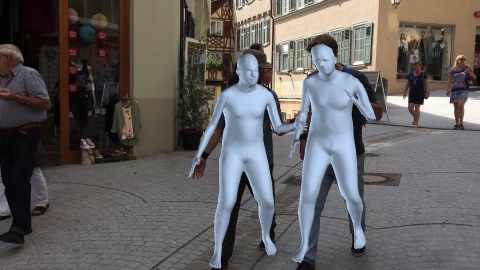}\\
  	\includegraphics[width=.25\textwidth]{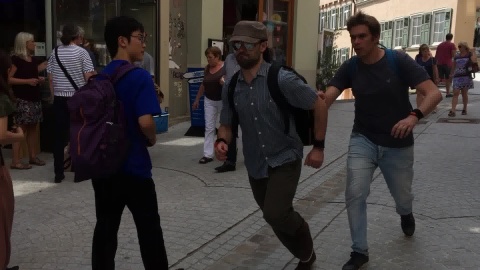}
    &\includegraphics[width=.25\textwidth]{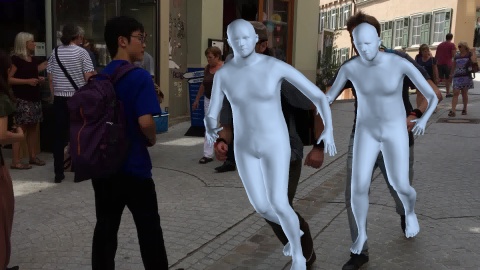}
  	&\includegraphics[width=.25\textwidth]{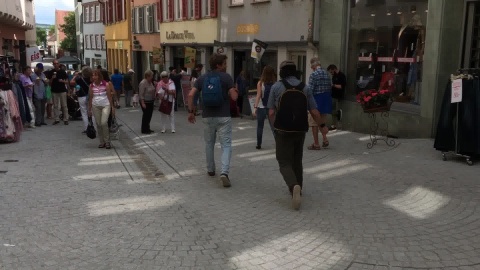}
    &\includegraphics[width=.25\textwidth]{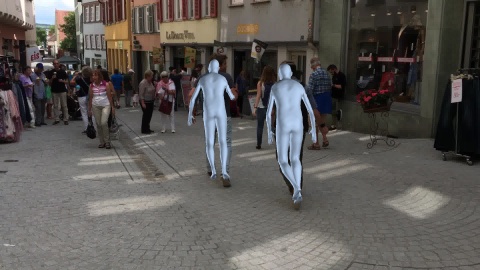}\\
 	Input frame & Our result & Input frame & Our result\\
    \end{tabular}
 \caption{Qualitative results of the proposed method on 3DPW sequence.}
\label{fig:visual} 
\end{figure*}

\section{Conclusion}\label{sec:conclusion}

We presented a novel nonparametric approach to reconstruct the 3D human mesh from a single image. Compared with the existing methods, our technique does not require any ground truth meshes during training. We introduced a Laplacian prior term and the part segmentation term in the loss function of the Graph CNN. In addition, we fed the pose estimation heatmaps and part segmentation masks to the feature embedding network to improve the robustness against occlusions. Experiments demonstrated that our technique is on par or outperforms existing techniques that use ground truth meshes in training.


{\small
\bibliographystyle{ieee}
\bibliography{npbody}
}

\end{document}